\documentclass[runningheads]{llncs}

\usepackage{cite}
\usepackage{amsmath,amssymb,amsfonts}
\usepackage{algorithmic}
\usepackage{graphicx}
\usepackage{hyperref}
\usepackage{textcomp}
\usepackage{xcolor}
\usepackage{float}
\usepackage{caption}
\usepackage{subcaption}
\usepackage{lineno}
\usepackage{booktabs}
\usepackage{enumitem}
\usepackage{multirow}
\usepackage{hyperref}

\makeatletter
\newcommand{\printfnsymbol}[1]{%
  \textsuperscript{\@fnsymbol{#1}}%
}
\makeatother

\usepackage{array}
\newcolumntype{L}[1]{>{\raggedright\let\newline\\\arraybackslash\hspace{0pt}}m{#1}}
\newcolumntype{C}[1]{>{\centering\let\newline\\\arraybackslash\hspace{0pt}}m{#1}}
\newcolumntype{R}[1]{>{\raggedleft\let\newline\\\arraybackslash\hspace{0pt}}m{#1}}

\begin{document}

\title{Anatomy of Domain Shift Impact on U-Net Layers in MRI Segmentation}

\titlerunning{Anatomy of Domain Shift in MRI Segmentation}

\author{
    Ivan Zakazov \thanks{equal contribution} \inst{1, 2} \and
    Boris Shirokikh \printfnsymbol{1} \inst{2} \and
    Alexey Chernyavskiy \inst{1} \and
    Mikhail Belyaev \inst{2}
}


\authorrunning{I. Zakazov, B. Shirokikh et al}

\institute{
    Philips Research, Moscow, Russia
    \and
    Skolkovo Institute of Science and Technology, Moscow, Russia
    \\
    \email{boris.shirokikh@skoltech.ru}
}

\maketitle


\begin{abstract}

Domain Adaptation (DA) methods are widely used in medical image segmentation tasks to tackle the problem of differently distributed train (source) and test (target) data. We consider the supervised DA task with a limited number of annotated samples from the target domain. It corresponds to one of the most relevant clinical setups: building a sufficiently accurate model on the minimum possible amount of annotated data. Existing methods mostly fine-tune specific layers of the pretrained Convolutional Neural Network (CNN). However, there is no consensus on which layers are better to fine-tune, e.g. the first layers for images with low-level domain shift or the deeper layers for images with high-level domain shift. To this end, we propose SpotTUnet -- a CNN architecture that automatically chooses the layers which should be optimally fine-tuned. More specifically, on the target domain, our method additionally learns the policy that indicates whether a specific layer should be fine-tuned or reused from the pretrained network. We show that our method performs at the same level as the best of the non-flexible fine-tuning methods even under the extreme scarcity of annotated data. Secondly, we show that SpotTUnet policy provides a layer-wise visualization of the domain shift impact on the network, which could be further used to develop robust domain generalization methods. In order to extensively evaluate SpotTUnet performance, we use a publicly available dataset of brain MR images (CC359), characterized by explicit domain shift. We release a reproducible experimental pipeline\footnote{\url{https://github.com/neuro-ml/domain_shift_anatomy}}.
\end{abstract}


\section{Introduction}
\label{sec:intro}


Whenever a model, trained on one distribution, is given some data, belonging to another distribution, a detrimental effect called \emph{domain shift} might pop up and decrease the inference quality \cite{wang2018deep}. This problem is especially acute in the field of medical imaging since any data collection instance (e.g., MRI apparatus) might appear to sample data, belonging to the domain of its own due to peculiarities of a model or the scanning protocol \cite{glocker2019machine}. Besides, preserving quality on the new data (i.e., \emph{domain adaptation}) is of utter importance because of the industry standards. A realistic set-up is that of \emph{supervised domain adaptation (sDA)}: some data from the new (\emph{target}) domain is labeled and should be utilized for fine-tuning the model, pre-trained on the \emph{source} domain.  

The central question of sDA research is \emph{how} should the net be fine-tuned. A great number of works adopt the transfer learning approach of fine-tuning the last layers only, which is underpinned by the notion of feature complexity increasing with depth \cite{yosinski2014transferable}. It is assumed, that low-level features should be shared across domains, while high-level ones are more prone to domain shift and should be therefore fine-tuned. However, in a number of Domain Adaptation papers the presence of low-level domain shift is demonstrated \cite{dou2018unsupervised, shirokikh2020first, zhao2021robust}. 

SpotTune \cite{guo2019spottune} is a \emph{Transfer Learning (TL)} approach, allowing for Adaptive Fine-tuning, providing, therefore, a domain shift stamp of each task by learning the corresponding fine-tuning policy. We employ SpotTune approach for getting better insight into the \emph{anatomy of domain shift} in the problem of Domain Adaptation.



\let\labelitemi\labelitemii
\noindent
Our contribution is threefold:
\begin{itemize}
    \item To the best of our knowledge, we are the first to propose SpotTUnet: SpotTune adapted for supervised DA in medical image segmentation
    \item We introduce interpretable regularization, with which SpotTune performs on par with the best of the alternative fine-tuning methods across the entire data availability spectrum 
    \item We study the optimal fine-tuning strategies for different data availability scenarios and provide intuition for the obtained results
\end{itemize}
    

\section{Related work}
\label{sec:related_work}


In this paper, we focus on the supervised DA setup in medical image segmentation. Our approach is motivated by SpotTune \cite{guo2019spottune}, which learns a policy to choose between fine-tuning and reusing pretrained network layers and at the same time fine-tunes chosen layers on the Target data (see details in Sec. \ref{sec:method}). Previous approaches in both supervised and unsupervised medical image DA mostly rely on the explicit choice of layers to fine-tune or to split the adversarial head from. We detail the most relevant approaches below.

The authors of \cite{yosinski2014transferable} have extensively evaluated features transferability in the image classification task. Their study suggests that features from the first layers of the network could be transferred between tasks, while the last layers should be re-trained. Contrary, convolutional filter reconstruction \cite{aljundi2016lightweight} is designed to tackle the low-level domain shift under unsupervised DA setup showing that the first layers are susceptible to domain shift more than the later ones. 


DA methods for medical image segmentation also tackle domain shift problem differently: the earlier approaches follow the motivation of \cite{yosinski2014transferable}, thus fine-tune the later layers of the network. According to the approach of \cite{kushibar2019supervised} fine-tuning of only the last CNN layer is performed, which yields improvement over transferring without adaptation. However, no comparison is provided with other supervised DA methods or various transferring strategies. Similarly, in \cite{valindria2018domain} the last CNN layer is fine-tuned, but the authors focus more on the training cases selection procedure rather than on the fine-tuning method development. Several works \cite{valverde2019one,ghafoorian2017transfer} provide a comparison between the outcomes of various numbers of the later layers being fine-tuned. Notably, in \cite{ghafoorian2017transfer} fine-tuning of the whole network is added to comparison, with better results demonstrated for a smaller number of the later layers fine-tuned. In the unsupervised DA setup, \cite{kamnitsas2017unsupervised} achieves better results adapting features from all the layers except for the first block, but the layers choice strategy remains unlearnable.

In contrast, later approaches follow the motivation of \cite{aljundi2016lightweight}, arguing that medical images (e.g. MRI) mostly contain low-level domain shift, which is due to varying intensity profiles but similar high-level structures, thus the first layers should be targeted. In \cite{shirokikh2020first} fine-tuning of the first layers is compared to fine-tuning of the last layers and of the whole network. The conclusion is that fine-tuning of the first layers is superior to fine-tuning of the last ones and is even preferable to fine-tuning of the whole network in the case of annotated Target data scarcity. In \cite{karani2021test}, an adaptive image normalization module is developed for the unsupervised test-time DA, which is conceptually close to fine-tuning of the first layers. Approaches of \cite{dou2018unsupervised,zhao2021robust} are also motivated by the same hypothesis and adapt the first layers in the unsupervised DA setup.

To this end, we compare SpotTUnet with the best unlearnable layer choice strategies within the supervised DA setup and show it to be a reliable tool for domain shift analysis. While authors of \cite{singh2020adaptation} demonstrate SpotTune to perform worse than histogram matching preprocessing in the medical image classification task, we argue that histogram matching is a task-specific method and show its extremely poor segmentation quality in our task. Many approaches competitive to SpotTune have been developed recently, but their focus is more narrow: obtaining the best score on the Target domain rather than analyzing domain shift properties. Therefore, we further study only the SpotTune-based approach.


\section{Method}
\label{sec:method}

The majority of supervised DA methods in medical image segmentation are based on the explicit choice of layers to fine-tune. However, as indicated in Sec.\ref{sec:related_work}, it is not always clear, whether the first or the last layers are to be targeted. Moreover, state-of-the-art architectures consist of skip-connections and residual paths \cite{bakas2018identifying}, while residual networks behave like ensembles of \textit{shallow} networks \cite{veit2016residual}. Therefore, it is also unclear which layers are actually the first across the most meaningful shallow sub-parts of the residual network.

\begin{figure}[h!]
\centering
\includegraphics[width=\textwidth]{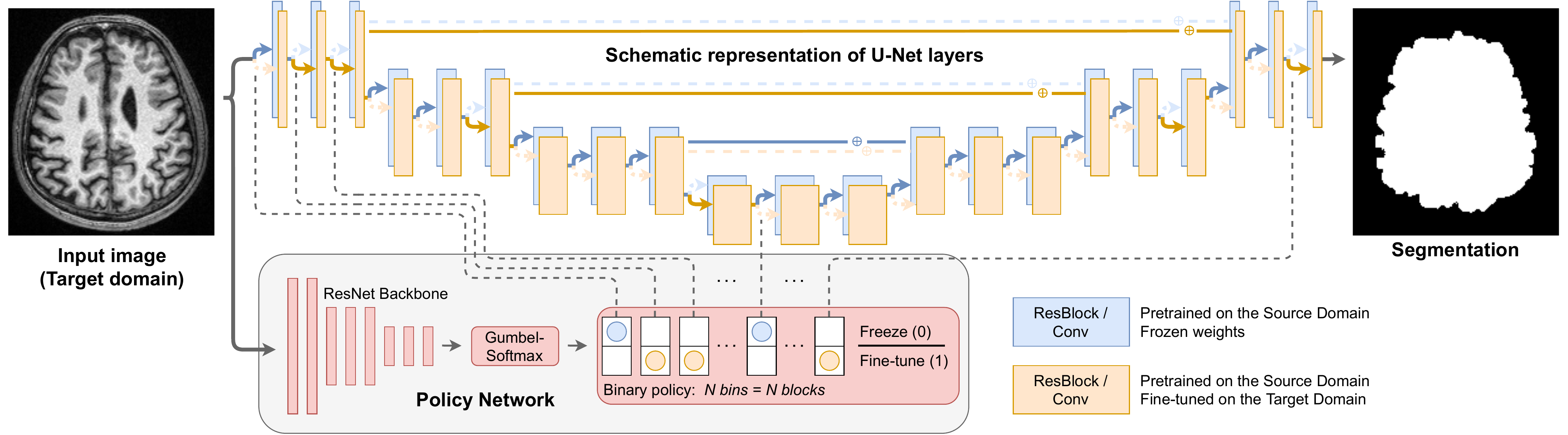}
\caption{SpotTUnet architecture for the supervised DA in medical image segmentation. We use U-Net architecture from \cite{shirokikh2020first} as the segmentation backbone, which is pretrained on the Source domain. The pretrained segmentation network is frozen (blue blocks) and has a copy (orange blocks) that is fine-tuned on the Target domain. The policy network is simultaneously trained on the Target domain to output binary decisions for each pair of blocks from the segmentation networks: use the frozen block (blue) \textit{vs} use the fine-tuned block (orange).}
\label{fig:spottune_seg}
\end{figure}

We introduce an extension of SpotTune \cite{guo2019spottune} on the supervised DA for medical image segmentation called \textbf{SpotTUnet}. SpotTUnet consists of two copies of the main (segmentation) network and a policy network (see Fig. \ref{fig:spottune_seg}). The main network is pretrained on the Source domain and then duplicated: the first copy has frozen weights (Fig. \ref{fig:spottune_seg}, blue blocks), while the second copy is fine-tuned on the Target domain (Fig. \ref{fig:spottune_seg}, orange blocks). The policy network predicts $N$ pairs of logits for each of $N$ segmentation network blocks (residual blocks or separate convolutions). For each pair of logits, we apply softmax and interpret the result as probabilities in a $2$-class classification task: class $0$ corresponds to the choice of a frozen block, while class $1$ means choosing to fine-tune the unfrozen copy. Then, for the $l$-th level of the network (frozen block is denoted $F_l$ and fine-tuned block $\tilde{F}_l$) we define its output as $\displaystyle x_l = I_l ( x ) F_l ( x_{l-1} ) + (1 - I_l ( x )) \tilde{F}_l ( x_{l-1} )$, where $I_l (x)$ is the indicator of choosing the frozen block (i.e. class $0$ probability  $> 0.5$). Here, we use Gumbel-Softmax to propagate the gradients through the binary indicator $I_l (x)$ exactly reproducing the methodology of SpotTune \cite{guo2019spottune}. Thus, we simultaneously train the policy network and fine-tune the duplicated layers.

Authors of SpotTune also propose a compact global policy (Global-k), which, via additional losses, constrains all the images to fine-tune the same $k$ blocks and aims at reducing the memory and computational costs while possibly decreasing the overall quality. We propose a simplified regularization approach that moreover yields higher quality in the cases of annotated data scarcity. Effectively, we apply $\mathbb{L}_1$ regularization term in the Global-$0$ SpotTune case, simultaneously minimizing the total number of fine-tuned blocks and achieving a more deterministic policy (exactly $0$ or $1$, which is due to $\mathbb{L}_1$ properties). The resulting loss is

\begin{equation}
    \mathcal{L} = \mathcal{L}_{segm} + \lambda \sum_{l=1}^N \left( 1 - I_l (x) \right),
\end{equation}

\noindent
where $\lambda$ is the balance parameter of the regularization term. Our motivation is different from the original: we assume that fewer blocks should be optimally fine-tuned in case of limited annotated data so that to avoid possible overfitting. Our regularization has only $1$ parameter compared to $3$ parameters of the original SpotTune regularization, and this parameter ($\lambda$) could be optimized during preliminary validation. The intuition behind $\lambda$ is simple: the less annotated data is available the larger $\lambda$ value should be.


\section{Experiments}
\label{sec:exp}

\subsection{Technical details}
\label{ssec:exp:tech}

\textbf{Data.} We report our results on a publicly available dataset CC359 \cite{souza2018open}. CC359 consists of $359$ MR images of head with the task being skull stripping. The dataset is split into $6$ equal domains which are shown \cite{shirokikh2020first} to contain domain shift resulting in a severe score deterioration. The data preprocessing steps are: interpolation to $1 \times 1 \times 1$ mm voxel spacing and scaling intensities into $0$ to $1$ interval. All splits and setups are detailed in Sec. \ref{ssec:exp:setup}.

\noindent 
\textbf{Metric.} To evaluate different approaches, we use surface Dice Score \cite{nikolov2018deep} at the tolerance of $1$ mm. While preserving the consistency with the methodology of \cite{shirokikh2020first}, we also find surface Dice Score to be a more representative metric for the brain segmentation task than the standard Dice Score.


\noindent
\textbf{Architecture and training.} The experimental evaluation provided in \cite{shirokikh2020first} shows that neither architecture nor training procedure variations, e.g. augmentation, affect the relative performance of conceptually different approaches. Therefore, in all our experiments we consistently use 2D U-Net architecture implementation from \cite{shirokikh2020first}. We also use a ResNet architecture \cite{he2016deep} as the policy network backbone. In all the experiments we minimize Binary Cross-Entropy loss ($\mathcal{L}_{segm}$) via stochastic gradient descent. Baseline and oracle (see Sec. \ref{ssec:exp:setup}) are trained for $100$ epochs ($100$ iterations per epoch) with the learning rate of $10^{-2}$ reduced to $10^{-3}$ at the $80$-th epoch. All fine-tuning methods are trained for $60$ epochs with the learning rate of $10^{-3}$ reduced to $10^{-4}$ at the $45$-th epoch. We ensure all the models reach the loss plateau. All models are trained with batch size $16$. The training takes about $4$ hours on a $16$GB nVidia Tesla V$100$ GPU \cite{zacharov2019zhores}.

\subsection{Experimental setup}
\label{ssec:exp:setup}

\paragraph{Baseline and oracle} The models trained on a single domain form the \textit{baseline} of our study. The transfer of such a model (without fine-tuning) on the other $5$ unseen domains results in quality deterioration, which is also shown in \cite{shirokikh2020first}. We also obtain scores within each domain (the \textit{oracle}) via $3$-fold cross-validation, thereby setting the upper bound for various DA methods.

\paragraph{SpotTUnet validation} Six domains yield $30$ Source-Target pairs, thus, $30$ supervised DA experiments. In order to avoid overfitting, we separate one Source domain (Siemens, $1.5$T) and, correspondingly, $5$ Source-Target pairs for SpotTUnet validation and use the other $25$ pairs for testing various DA approaches. On the $5$ validation pairs, we firstly adjust the temperature parameter of Gumbel-Softmax ($\tau$) via grid-search over $\tau \in \{ .01, .1 , .5, 1, 2, 5 \}$. The number of annotated slices from the Target domain, in this case, is $270$ (one 3D image). Secondly, we search for the optimal $\lambda$ for each amount of annotated Target data considered via grid-search over $\lambda \in \{0, 1, 3 , 5, 7, 10, 12, 15, 20 \; (\times 10^{-3}) \}$. In both validation and testing experiments, we study the same setups of the annotated Target data scarcity: $8$, $12$, $24$, $45$, $90$, $270$, and $800$ slices available for fine-tuning. The optimal value of $\lambda$ is fixed for each data scarcity setup and used for SpotTUnet when testing on the remaining $25$ pairs.

\paragraph{Supervised DA methods} On the rest of the $25$ testing pairs, we compare $4$ methods: \textit{fine-tuning of the first network layers}, \textit{fine-tuning of the whole network} from \cite{shirokikh2020first}, \textit{histogram matching} from \cite{singh2020adaptation}, and SpotTUnet. We load a \textit{baseline} model pretrained on the corresponding Source domain and then fine-tune it via one of the methods or preprocess the Target data in case of histogram matching. We compare methods by averaging surface Dice Scores over the Target images, separated for test (omitted when fine-tuning).

\subsection{Results and discussion}
\label{ssec:exp:results}

\begin{figure}[h!]
\centering
\includegraphics[width=\textwidth]{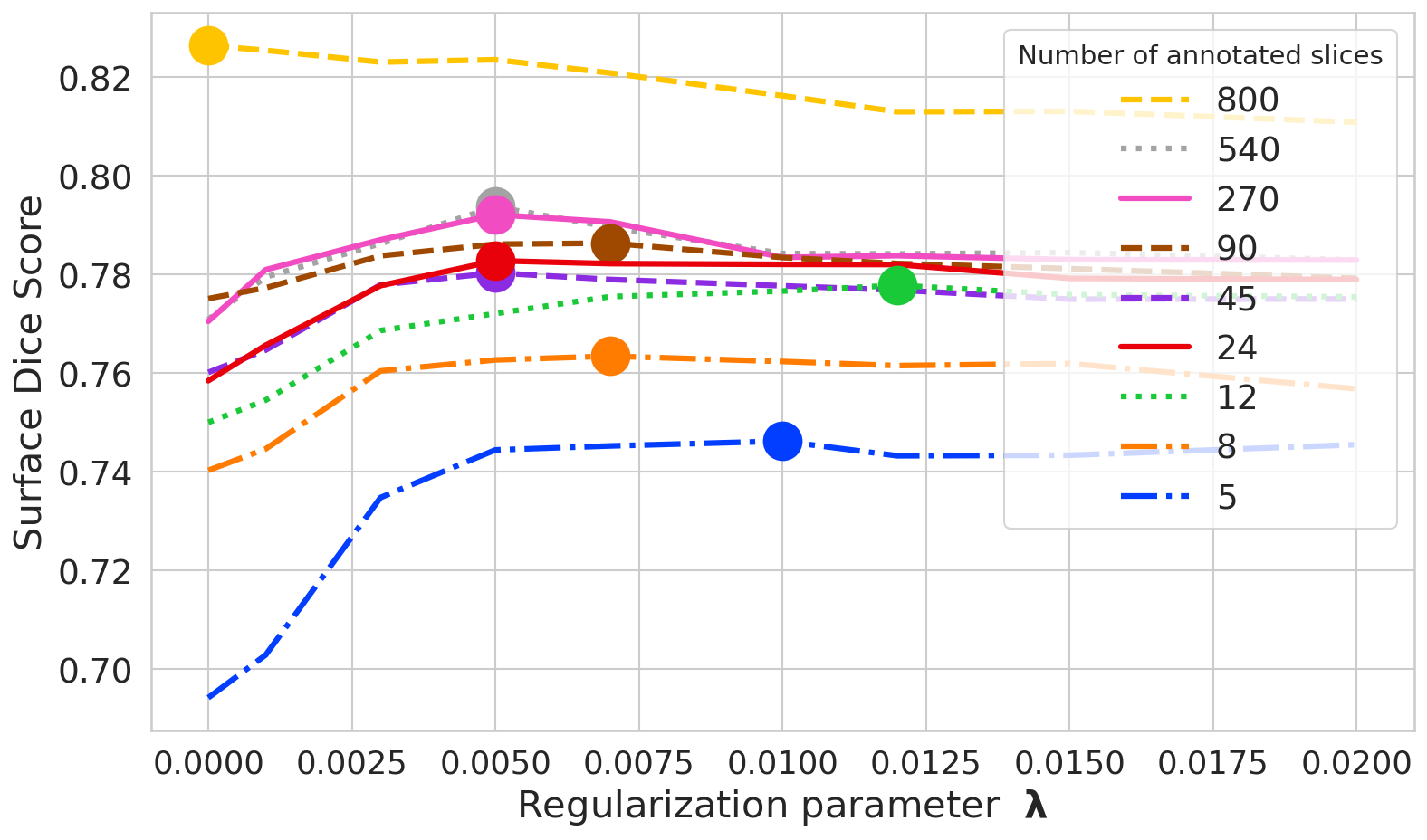}
\caption{The validation performance of SpoTUnet dependence on the regularization parameter $\lambda$. In each point, the average surface Dice Score over 5 validation experiments is calculated. Each line corresponds to some amount of available annotated data from the Target domain. The bold points indicate the optimal $\lambda$ values in terms of surface Dice Score.}
\label{fig:lambda}
\end{figure}

We firstly find and fix hyperparameters for SpotTUnet through validation: the temperature of Gumbel-Softmax is set to $\tau = 0.1$ and the optimal regularization $\lambda$ is set to the optimal value for each data scarcity setup (see Fig. \ref{fig:lambda}). We also note that Gumbel-Softmax training stability is extremely sensitive to $\tau$ choice, thus suggest to validate different values of $\tau$ at one of the first stages of deploying SpotTune-like architectures. Positive regularization term benefits almost all data scarcity setups: surface Dice Score of the optimal $\lambda$ is significantly higher ($p < 10^{-3}$, one-sided Wilcoxon signed-rank test) than the surface Dice Score of the corresponding model without the regularization. The only exception is the case of $800$ available Target slices (or three $3$D images), where the optimal $\lambda$ is close to $0$, and the quality drops with the increase of $\lambda$ (see Fig. \ref{fig:lambda}). We conclude that while SpotTUnet learns the optimal policy without regularization when there is no Target data shortage, regularization improves DA performance significantly in case of Target data scarcity.

\begin{figure}[h!]
\centering
\includegraphics[width=\textwidth]{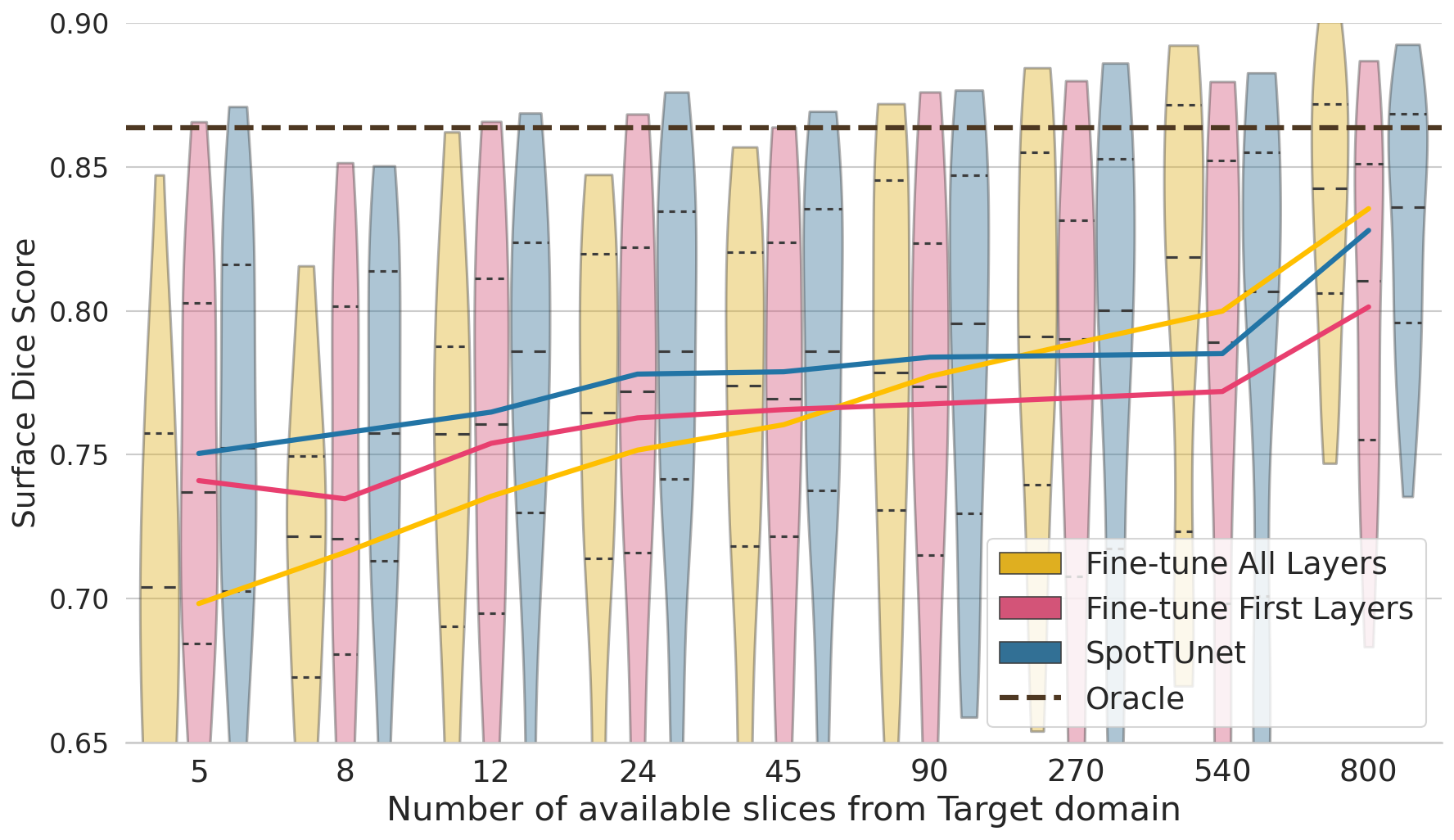}
\caption{The methods performance dependence on the amount of available Target data. Baseline and histogram matching yield poor quality, thus not included.}
\label{fig:sdcs}
\end{figure}

We further compare SpotTUnet with \emph{Fine-Tuning All Layers}, \emph{Fine-Tuning the First Layers}, and \textit{histogram matching}. We present both distributions of the surface Dice Score (violin plots) and their average values (corresponding lines) in Fig. \ref{fig:sdcs}. Histogram matching achieves only $0.29$ average surface Dice Score, which is even lower than the \textit{baseline} average score of $0.55$; we exclude both methods from the comparison in Fig. \ref{fig:sdcs}. Here, we show that SpotTune performs at the same level as the best of the other methods regardless of the Target data scarcity severity.


\begin{figure}[h!]
\centering
\includegraphics[width=\textwidth]{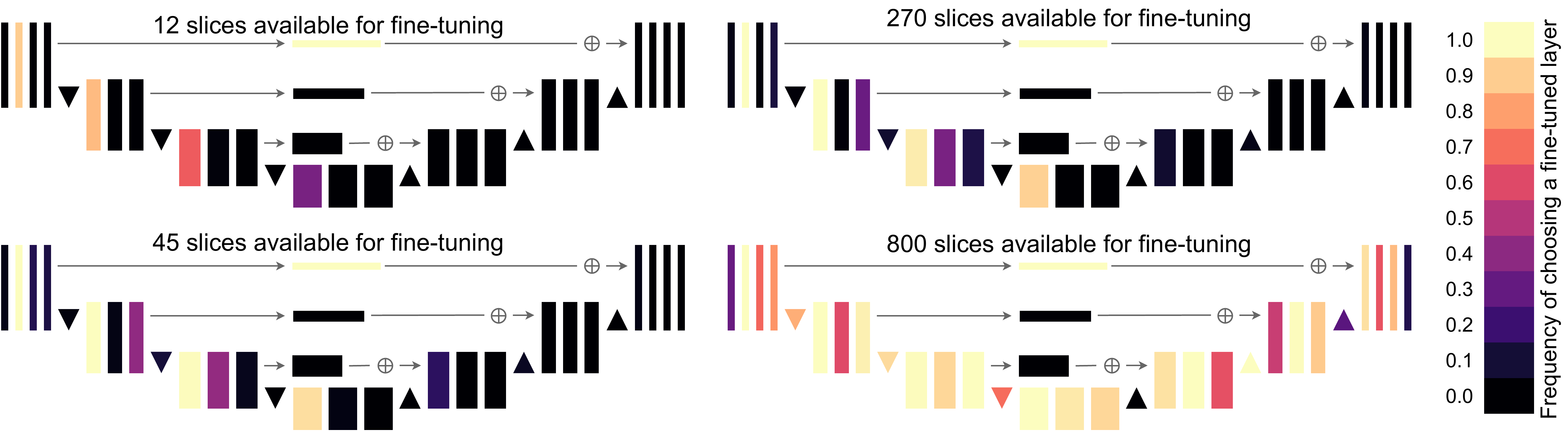}
\caption{SpotTUnet learnt policy visualization for the cases of $12$ (upper-left), $45$ (bottom-left), $270$ (upper-right), and $800$ (bottom-right) available Target slices. Colored blocks correspond to either residual blocks or convolutions. Triangular blocks are the convolutions that perform $\times 2$ up- or down-sampling.}
\label{fig:layerswise_template}
\end{figure}

Finally, we track SpotTUnet policy on the test data: for each layer, we calculate the frequency of choosing the fine-tuned one instead of the pretrained and frozen one. The layer-wise visualization is given in Fig. \ref{fig:layerswise_template}. We find that blocks from the encoder part of U-Net are more likely to be fine-tuned, especially in the case of annotated data scarcity. However, these are not exactly the first layers (e.g., preserving the original resolution), which opposes the conclusion of \cite{shirokikh2020first}. We further hypothesize, that SpotTUnet policy indicates layers that should be fine-tuned for the optimal solution. Consequently, feature maps preceding these frequently fine-tuned layers  might be marked with drastic domain shift. We note that it is worth evaluating if unsupervised DA approaches \cite{kamnitsas2017unsupervised,zhao2021robust} would benefit from passing these SpotTUnet indicated domain shift reach feature maps to the adversarial heads and leave this hypothesis validation for the future research.

We attribute the difference between the policies observed and those presented in the original SpotTune paper \cite{guo2019spottune} (mostly final layers fine-tuned) to the fundamental difference between Transfer Learning (TL) and DA. In TL one deals with data of varying nature, thus the later layers should be addressed; in DA, the datasets contain semantically homogeneous data (e.g., brain MRI scans), thus domain shift is mostly low-level and the first layers should be targeted.

\section{Conclusion}
\label{sec:conclusion}


We propose a fine-tuning approach for supervised DA in medical image segmentation called SpotTUnet. Our experiments demonstrate SpotTUnet to preserve the quality of the alternative methods while eliminating the need for switching between various methods depending on the Target data availability. Besides, it learns automatically, which layers are to be optimally fine-tuned on the target domain, therefore providing a policy, indicative of the network layers most susceptible to domain shift. We believe that SpotTUnet generated policy might be used for developing more robust unsupervised DA methods, which is the goal of our future research.

\bibliographystyle{splncs04}
\bibliography{main.bib}

\end{document}